\documentclass[11pt,a4paper]{article}
\usepackage[hyperref]{emnlp2020-templates/emnlp2020}
\usepackage{times}
\usepackage{latexsym}

\usepackage{microtype}

\usepackage{booktabs} 
\usepackage{amsmath}

\usepackage{amsfonts}
\usepackage{bm}
\usepackage{array}
\usepackage{enumitem}
\usepackage{tikz}
\usepackage{pgfplots}
\usepackage{multirow}
\usepackage{makecell}
\usepackage{xcolor}
\usepackage{amssymb}
\usepackage{nicefrac}       
\usepackage{microtype}      
\usepackage{url}

\newcommand\setrow[1]{\gdef\rowmac{#1}#1\ignorespaces}
\newcommand\clearrow{\global\let\rowmac\relax}

\clearrow
\newcommand{\wideslash}{\text{ / }}

\newcommand{\checkhere}[1]{{#1}}
\newcommand*{\rom}[1]{\romannumeral#1\relax}
\newcommand{\centercell}[1]{\multicolumn{1}{c}{#1}}

\aclfinalcopy 
\def\aclpaperid{690} 

\title{Group-wise Contrastive Learning for Neural Dialogue Generation}

\author{
Hengyi Cai$^{\dagger,\S}$\thanks{\ \ Work done at JD.com.}, Hongshen Chen$^\ddagger$\\
{\bf Yonghao Song$^\dagger$, Zhuoye Ding$^\ddagger$, Yongjun Bao$^\ddagger$, Weipeng Yan$^\ddagger$, Xiaofang Zhao$^{\dagger}$} \\
$^\dagger${Institute of Computing Technology, Chinese Academy of Sciences, Beijing, China} \\
$^\S${University of Chinese Academy of Sciences, Beijing, China} \\
$^\ddagger${JD.com, China} \\
{caihengyi@ict.ac.cn, ac@chenhongshen.com} \\
{\{songyonghao, zhaoxf\}@ict.ac.cn, \{dingzhuoye, baoyongjun, Paul.yan\}@jd.com}
}

\date{}

\begin{document}
\maketitle

\begin{abstract}
\checkhere{
Neural dialogue response generation has gained much popularity in recent years. 
Maximum Likelihood Estimation (MLE) objective is widely adopted in existing dialogue model learning.
However, models trained with MLE objective function are plagued by the low-diversity issue when it comes to the open-domain conversational setting.
Inspired by the observation that humans not only learn from the positive signals but also benefit from correcting behaviors of undesirable actions, in this work, we introduce contrastive learning into dialogue generation, where the model explicitly perceives the difference between the well-chosen positive and negative utterances.
Specifically, we employ a pretrained baseline model as a reference.
During contrastive learning, the target dialogue model is trained to give higher conditional probabilities for the positive samples, and lower conditional probabilities for those negative samples, compared to the reference model.
To manage the multi-mapping relations prevalent in human conversation, we augment contrastive dialogue learning with group-wise dual sampling.
Extensive experimental results show that the proposed group-wise contrastive learning framework is suited for training a wide range of neural dialogue generation models with very favorable performance over the baseline training approaches.
}
\end{abstract}

\section{Introduction}

 Open-domain human-machine dialogue systems, especially the generation-based genre, have attracted extensive attention recently.
 Typically, following the neural encoder-decoder paradigm, contemporary dialogue generation models~\citep{DBLP:conf/acl/ShangLL15,DBLP:conf/aaai/SerbanSBCP16,DBLP:conf/aaai/XingWWLHZM17,ijcai2018-778,10.1145/3383123,Liu2020YouIM}, more often than not,
 are trained with Maximum Likelihood Estimation (MLE) principle to mimic human context-response pairs in the training corpus.
 While notable gains have been achieved under this learning framework, prior art~\citep{DBLP:conf/naacl/LiGBGD16,DBLP:conf/emnlp/LiMSJRJ17,DBLP:conf/ijcai/ZhangLGXC18} suggests that naive MLE objective used for training neural dialogue generation models is not that effective enough and tends to result in issues like dull response generation.
 By optimizing the likelihood of training dialogues, neural models are inclined to assign high probabilities to ``safe'' responses, due to the fact that vacuous responses like ``I don't know'' are of relatively high frequencies in conversational datasets~\citep{DBLP:conf/naacl/LiGBGD16}.

\begin{figure*}[!ht]
    \centering 
    \includegraphics[width=1.0\textwidth]{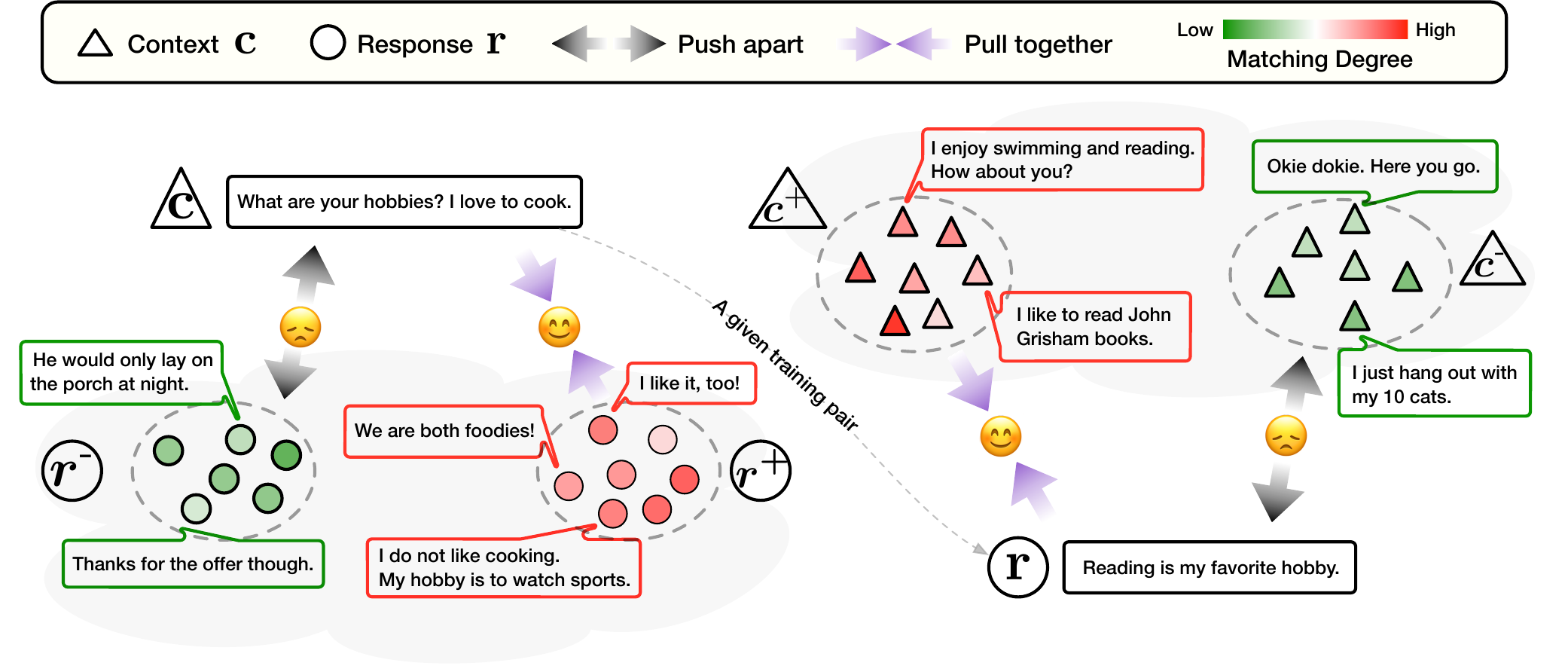}
    \caption{
        An illustration case of group-wise contrastive learning.
        For a given training instance, the proposed framework explicitly considers the multi-mapping relations in human conversations, by encouraging the dialogue generation model to pull the matched sample pairs together and push the mismatched pairs apart in the latent space.
    }
    \label{fig:motivation}
\end{figure*}

 One promising training framework for neural dialogue generation is adversarial learning~\citep{NIPS2014_5423,DBLP:conf/emnlp/LiMSJRJ17}, where a discriminator provides rewards for the generator by contrastively distinguishing dialogues as human-generated or machine-generated. 
 However, the learning ability of GANs in text is drastically limited due to training instability and model collapse~\citep{DBLP:conf/iclr/NieNP19,DBLP:conf/iclr/CacciaCFLPC20}.
 \textit{First}, the discriminator is usually unlikely to be fooled very easily, and the generator can hardly learn from those ineffective rewards.
 \textit{Second}, the generator is sometimes encouraged to mimic the high-frequency generic responses in the training corpus, \checkhere{because in some cases, the discriminator fails to distinguish a good response from a bad one: it can easily recognize contentful but less-grammatical responses as machine-generated, yet treat those human-generated dull responses as the oracle.}
 
 In this paper, we introduce contrastive learning~\citep{DBLP:conf/cvpr/HadsellCL06,DBLP:journals/jmlr/GutmannH12} into dialogue generation, where the model explicitly perceives the difference between the well-chosen positive and negative utterances.
 From the perspective of contrastive learning, the discriminator in adversarial learning considers human-generated responses as positive utterances and synthetic ones as negative samples.
 Instead, this work deems highly-matched context-response pairs as positive samples and mismatched training pairs as negative samples.
 In particular, we utilize a pretrained baseline model as a reference.
 During contrastive learning, for context $\bm{c}$ and its response $\bm{r}$, the target dialogue model is trained to give higher conditional probabilities $p(\bm{r}|\bm{c})$ for the positive samples, and lower conditional probabilities for those negative samples, compared to the reference model.
 This training paradigm encourages the model to pull the positive data points together and push apart the negative samples, as exemplified in Figure~\ref{fig:motivation}.
 As a result, our proposed training scheme explicitly takes the semantic associations and differences among training examples into account for dialogue modeling.
 Besides, by contrastively characterizing the distinctions relative to a strong reference, our method implicitly enhances the {distinctiveness} of the generated responses as well, and ensures that the overall performance of the target model is not inferior to the reference.
 
 Contrastively learning from one pair of positive and negative samples is quite straightforward, however, multi-mapping relations prevail in human-human conversations, where there exist multiple appropriate responses for a given context, and a response sometimes fits well to several contexts, known as one-to-many and many-to-one relations.
 Such complex multi-mapping relations are overlooked in previous learning framework, which hampers effective dialogue response learning.
 Furthermore, if a potential highly-matched utterance pair is treated as the negative sample or an outlier is used as the positive sample, the model may be confused. 
 Therefore, in order to consider the multi-mapping phenomenon in human conversations and remedy the potential problematic false learning samples, and enhance the training stability, we augment contrastive learning with group-wise dual sampling, where groups of positive and negative instances are sampled regarding both the context and the response, respectively.
 To further depict subtle differences between instances in the group, we adapt the instance importance with the matching scores, and optimize the weighted loss.
 
  We show an illustration case to understand our learning framework in Figure~\ref{fig:motivation}.
  Given a training context-response pair $(\bm{c}, \bm{r})$, for context ``What are your hobbies? I love to cook'', multiple highly-matched responses are organized as the positive samples $\bm{r}^\text{+}$, and the mismatched utterances are deemed as the negatives $\bm{r}^\text{-}$.
  On the dual direction, regarding the response ``Reading is my favorite hobby'', multiple sampled context utterances are similarly divided into $\bm{c}^\text{+}$ and $\bm{c}^\text{-}$.
  Compared with the reference baseline, the target dialogue model is trained to give higher generation probabilities for positive instances, $(\bm{c}, \bm{r}^\text{+})$ and $(\bm{c}^\text{+}, \bm{r})$, and lower probabilities for negatives $(\bm{c}, \bm{r}^\text{-})$ and $(\bm{c}^\text{-}, \bm{r})$.
  By this mean, the target model is actually induced to pull the positive sample pairs together and push the mismatched pairs apart, and thus learns from the distinctions between the positives and negatives.

 The proposed group-wise contrastive learning framework is suited for training various neural dialogue generation models.
 We conduct extensive studies on three large-scale conversation datasets using four popular dialogue models to assess the proposed approach.
 The experimental results confirm the effectiveness of our learning framework with very favorable performance over the baseline training approaches\footnote{Code is available at \url{https://github.com/hengyicai/ContrastiveLearning4Dialogue}}.
\begin{figure*}[!t]
    \centering 
    \includegraphics[width=1.0\textwidth]{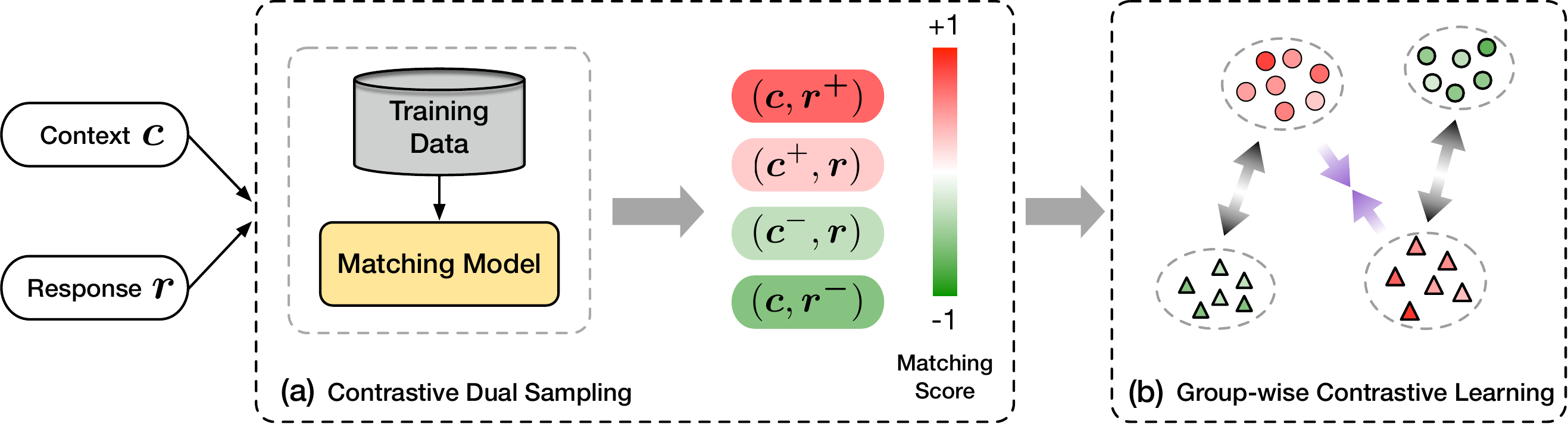}
    \caption{
        A demonstration of the proposed group-wise contrastive dialogue learning pipeline.
        For each training pair, it first samples a group of highly-matched examples and another group of most mismatched utterances regarding both the context and response to build the contrastive samples, using an off-the-shelf conversation matching model (\S\ref{sec:Contrastive_Dual_Sampling}).
        The target dialogue model is then trained with group-wise contrastive learning (\S\ref{sec:GCL}).
    }
    \label{fig:model_overview}
\end{figure*}

\section{Contrastive Dialogue Learning}

\subsection{Dialogue Learning by Comparison}

Given training data $\mathbb{D}$ containing context-response pairs $\{(\bm{c}, \bm{r})_i\}_{i=1}^{N}$, a dialogue model parameterized by ${\bm{\theta}}$ aims to map from the input context $\bm{c}$ to the output response $\bm{r}$.
To achieve this, conventional dialogue learning approaches search the parameter ${\bm{\theta}}$ by maximizing the conditional probability $p_{\bm{\theta}}(\bm{r}|\bm{c})$ over the training samples.
MLE maximizes the log-likelihood of training pairs while adversarial based approaches rely on the discriminator to distinguish between good responses and bad ones.
To combat the aforementioned drawbacks of traditional training approaches in dialogue learning, we advocate the use of contrastive learning to explicitly perceive the difference between the positive and negative samples.
Inspired by ~\citet{DBLP:journals/jmlr/GutmannH12,CL4ImageCaption}, we utilize a pretrained baseline model $p_n(\cdot;{\bm{\phi}})$, to provide the target dialogue model $p_m(\cdot;{\bm{\theta}})$ a strong reference when contrasting the positive samples and the negatives.
Humans not only learn from the positive signals but also benefit from correcting behaviors of undesirable actions.
Intuitively, the target dialogue model is expected to give higher conditional probabilities $p(\bm{r}|\bm{c})$ for the positive samples, and lower conditional probabilities for those negative samples, compared to the reference model.
Towards this end, we define the difference between $p_m(\bm{r}|\bm{c}, \bm{\theta})$ and $p_n(\bm{r}|\bm{c},\bm{\phi})$ as:
\begin{equation}
    \small
    \begin{split}
        \mathcal{D}((\bm{c}, \bm{r}); \bm{\theta}, \bm{\phi}) =& \log{\frac{p_m(\bm{r}|\bm{c}, \bm{\theta})}{p_n(\bm{r}|\bm{c}, \bm{\phi})}}.
    \end{split}
\end{equation}
\normalsize
We wish that $\mathcal{D}((\bm{c}, \bm{r}); \bm{\theta}, \bm{\phi}) > 0$ for any positive pair and vice versa for any negative pair.
Concretely speaking, we minimize the following loss function:
\begin{equation}
    \small
    \begin{split}
    &\mathcal{L}''(\bm{\theta}; \mathbb{D}, \bm{\phi})= \\
    & -\frac{1}{N}\sum_{(\bm{c}, \bm{r})\in\mathbb{D}}\log{\sigma(\mathcal{D}((\bm{c}, \bm{r})^\text{+}; \bm{\theta}, \bm{\phi}))} \\
    & -\frac{1}{N}\sum_{(\bm{c}, \bm{r})\in\mathbb{D}}\log{[1-\sigma(\mathcal{D}((\bm{c}, \bm{r})^\text{-}; \bm{\theta}, \bm{\phi}))]}
    \end{split},
    \label{eq:loss''}
\end{equation}
\normalsize
where $\sigma(\cdot)$ is the sigmoid activation function, the given training pair $(\bm{c}, \bm{r})$ can be used as the positive sample $(\bm{c}, \bm{r})^\text{+}$ and the negative sample $(\bm{c}, \bm{r})^\text{-}$ can be obtained through negative sampling using the given instance $(\bm{c}, \bm{r})$.

Optimizing the dialogue model with the above objective function is reminiscent of nonlinear logistic regression in Noise-Contrastive Estimation (NCE)~\citep{DBLP:journals/jmlr/GutmannH12}.
The underlying motivation of our formulation and NCE are essentially different.
The reference model in our work is utilized to constrain the behaviors of the target model, rather than serve as a noise distribution to provide noise data.
Another difference is that, instead of using the log-ratio between $p_m(\cdot;\bm{\theta})$ and $p_n(\cdot;\bm{\phi})$ to compute posterior classification probabilities as in NCE, we introduce the function $\mathcal{D}((\bm{c}, \bm{r}); \bm{\theta}, \bm{\phi})$ to characterize the distinctions of intrinsic dialogue properties relative to the reference, and encourage the generation of positive samples as well as penalize the negative ones through minimizing the loss in Eq.(\ref{eq:loss''}).
Besides, by contrastively characterizing the distinctions relative to a strong reference, our method implicitly enhances the {distinctiveness} of the generated response as well, and ensures that the overall performance of the target model is not inferior to the reference.

\subsection{Contrastive Dual Sampling}
\label{sec:Contrastive_Dual_Sampling}

Nevertheless, in the presence of multi-mapping relations in human dialogues, effectively sampling the positive and negative pairs in conversation is not that straightforward and even runs the risk of introducing false learning samples.
To manage the complex multi-mapping phenomenon in human conversations and enhance the training stability, we augment the contrastive learning with group-wise dual sampling, where groups of positive and negative instances are sampled regarding both the context and the response, respectively.
To put it concretely, for each training instance $(\bm{c}, \bm{r})$, we find a group of positive examples $\{(\bm{c}, \bm{r^\text{+}})_i\}_{i=1}^{\bm{k}}$ with highest matching degree and a group of negative examples $\{(\bm{c}, \bm{r^\text{-}})_i\}_{i=1}^{\bm{k}}$ with lowest matching degree, using an off-the-shelf pretrained matching model to compute the matching scores between the given context and candidate responses.
Similarly, $\{(\bm{c^\text{+}}, \bm{r})_i\}_{i=1}^{\bm{k}}$ and $\{(\bm{c^\text{-}}, \bm{r})_i\}_{i=1}^{\bm{k}}$ are also retrieved from the training set to serve as the context-side contrastive examples, as shown in Figure~\ref{fig:model_overview}(a).
In this work, we adopt MSN~\citep{yuan_multi-hop_2019}, a context-response matching network based on multi-hop selection, as the off-the-shelf matching model.
Note that other sophisticated matching models can also be applied, e.g., deep attention matching network~\citep{zhou_multi-turn_2018}.

\subsection{Group-wise Contrastive Learning}
\label{sec:GCL}
For each training instance $(\bm{c}, \bm{r})$, as describe in \S\ref{sec:Contrastive_Dual_Sampling}, we sample $\bm{k}$ different positive and negative pairs regarding both the dialogue context and its response to manage multi-mapping relations in conversation and stabilize the model training.
The resultant well-chosen samples are composed of positive samples, {$\{(\bm{c}, \bm{r^\text{+}})_i\}_{i=1}^{\bm{k}}$} and {$\{(\bm{c^\text{+}}, \bm{r})_i\}_{i=1}^{\bm{k}}$}, and the negatives, {$\{(\bm{c}, \bm{r^\text{-}})_i\}_{i=1}^{\bm{k}}$} and {$\{(\bm{c^\text{-}}, \bm{r})_i\}_{i=1}^{\bm{k}}$}.
Then, the loss function is updated as:
\begin{equation}
    \small
    \begin{split}
    &\mathcal{L}'(\bm{\theta}; \mathbb{D}, \bm{\phi})= \\
    & -\frac{1}{N}\sum_{(\bm{c}, \bm{r})\in\mathbb{D}}\frac{1}{2\bm{k} + 1}\sum_{i=1}^{2\bm{k} + 1}\log{\sigma(\mathcal{D}((\bm{c}, \bm{r})_i^\text{+}; \bm{\theta}, \bm{\phi}))} \\
    & -\frac{1}{N}\sum_{(\bm{c}, \bm{r})\in\mathbb{D}}\frac{1}{2\bm{k}}\sum_{i=1}^{2\bm{k}}\log{[1-\sigma(\mathcal{D}((\bm{c}, \bm{r})_i^\text{-}; \bm{\theta}, \bm{\phi}))]}
    \end{split}.
    \label{eq:loss'}
\end{equation}
\normalsize

Given varied matching degrees of the collected context-response pairs in open-domain dialogue,
indiscriminately training on such data impedes the model to perceive intra-group differences of these samples.
We thus utilize the matching score $\bm{s}$ attached with each sample to adapt its instance effect on the group-wise contrastive dialogue learning.
Specifically, for a given training example $(\bm{c}, \bm{r})$, the matching score $\bm{s}^\text{+}$ of its positive pair lies in $(0, 1]$ and the negative score $\bm{s}^{\text{-}}$ lies in $[-1, 0]$.
To induce the model learning from sample pairs with varied matching degrees discriminately, the loss function is finally defined to be:
\begin{equation}
    \small
    \begin{split}
    &\mathcal{L}(\bm{\theta}; \mathbb{D}, \bm{\phi})= \\
    & -\frac{1}{N}\sum_{(\bm{c}, \bm{r})\in\mathbb{D}}\frac{1}{2\bm{k} + 1}\sum_{i=1}^{2\bm{k} + 1}\log{[\bm{s}_i^\text{+}\cdot\sigma(\mathcal{D}((\bm{c}, \bm{r})_i^\text{+}; \bm{\theta}, \bm{\phi}))]} \\
    & -\frac{1}{N}\sum_{(\bm{c}, \bm{r})\in\mathbb{D}}\frac{1}{2\bm{k}}\sum_{i=1}^{2\bm{k}}\log{[1+\bm{s}_i^\text{-}\cdot\sigma(\mathcal{D}((\bm{c}, \bm{r})_i^\text{-}; \bm{\theta}, \bm{\phi}))]}
    \end{split}.
    \label{eq:loss_with_matching_score}
\end{equation}
\normalsize
The loss function $\mathcal{L}(\bm{\theta})$ reaches its lower bound when the positive and negative pairs can be perfectly distinguished,
i.e., $p_m(\bm{r}|\bm{c}, \bm{\theta}) \gg p_n(\bm{r}|\bm{c}, \bm{\phi})$ for the positive samples and $p_m(\bm{r}|\bm{c}, \bm{\theta}) \ll p_n(\bm{r}|\bm{c}, \bm{\phi})$ for the negatives, which indicates that the target dialogue model is able to clearly contrast a group of positive candidates from the negative ones and generate highly-distinctive responses for the given contexts.

\subsection{Discussion}
Neural sequence-to-sequence models trained with the MLE objective function are plagued by the low-diversity issue when it comes to the open-domain conversational setting, in which bland and generic utterances usually dominate the data distribution.
Since the objective of MLE is to maximize only the probabilities of ground-truth context-response pairs, it fails to capture the multi-mapping nature of human dialogues, not to mention the semantic differences among various candidates for a given example.
While the proposed group-wise contrastive learning framework explicitly explores multiple variants of a given dialogue example by leveraging an off-the-shelf matching model, and implicitly guarantees the ground-truth generation probabilities through the contrastive constraints in Eq.(\ref{eq:loss_with_matching_score}).

\checkhere{Adversarial learning approaches and our proposed framework both involve an auxiliary model during the training process.
However, GANs are learned via a competition between the target generator and the counteracting discriminator, which needs careful tuning to prevent model collapse in text modeling~\citep{DBLP:conf/iclr/CacciaCFLPC20},
whereas in our framework, the auxiliary reference system models conversation data in the same direction with the target dialogue model, and is stable during the learning procedure.}

\begin{table}[!t]
    \centering
    \resizebox{1.0\columnwidth}{!}{
        \begin{tabular}{@{}lrrr@{}}
        \toprule
        & PersonaChat & Douban & OpenSubtitles \\
        \midrule
        \#context-response pairs          & 140,248 & 218,039 & 353,046   \\
        Avg. \#turns per context          & 2.69    & 3.94    & 3.79     \\
        Avg. \#words per utterance        & 11.96   & 15.28   & 6.85    \\
        Training Pairs         & 113,558 & 198,039 & 293,129 \\
        Validation Pairs       & 13,602  & 10,000  & 29,960  \\
        Test Pairs             & 13,088  & 10,000  & 29,957  \\
        \#Tokens               & 18,029  & 40,000  & 40,000 \\
        \bottomrule
        \end{tabular}
    }
    \caption{
        Data statistics for PersonaChat, Douban and OpenSubtitles datasets.
    }
    \label{tbl:data_statics}
\end{table}

\begin{table*}[!ht]
\centering
\resizebox{1.0\textwidth}{!}{
\centering
\begin{tabular}{p{0.04\columnwidth}|>{\rowmac}l>{\rowmac}r>{\rowmac}r>{\rowmac}r>{\rowmac}r>{\rowmac}r>{\rowmac}r>{\rowmac}r>{\rowmac}r>{\rowmac}r<{\clearrow}}
\toprule
 & Models & \centercell{BLEU-1\wideslash2\wideslash3\wideslash4} & \centercell{Dist-1} & \centercell{Dist-2} & \centercell{Dist-3} & \centercell{Avg.} & \centercell{Ext.} & \centercell{Gre.} & \centercell{Coh.} & \centercell{Ent-1\wideslash2} \\
 \midrule
 \multirow{8}{*}{(a)} 
 & \textsc{Seq2Seq} & 12.040\wideslash3.9950\wideslash0.8815\wideslash0.2312 & 0.4309 & 2.045 & 4.303  & 36.33 & 28.66 & 63.64 & 41.66 & 6.891\wideslash10.81 \\
 \setrow{\bfseries} 
 &\normalfont{\textsc{Seq2Seq}} (${\blacktriangleright}$) & 13.660\wideslash4.9160\wideslash1.5970\wideslash0.6122 & 0.8492 & 5.093 & 12.000 & 39.76 & 31.74 & 64.76 & 49.39 & 7.016\wideslash10.90 \\ \cline{2-11}
 &{HRED} & 12.410\wideslash3.8360\wideslash0.8455\wideslash0.2364 & 0.4744 & 2.546 & 6.127 & 36.52 & 28.37 & 64.12 & 39.08 & 6.792\wideslash10.66 \\
 \setrow{\bfseries} 
 &\normalfont{HRED} (${\blacktriangleright}$) & 13.180\wideslash4.3220\wideslash1.0360\wideslash0.3274 & 0.7130 & 4.468 & 11.220 & 38.54 & 29.65 & 64.26 & 44.07 & 6.931\wideslash10.84 \\ \cline{2-11}
 &\textsc{Transformer} & 11.460\wideslash3.2080\wideslash0.5389\wideslash0.1476 & 0.4813 & 2.544 & 6.146 & 35.72 & 27.38 & 63.61 & 38.02 & 6.804\wideslash10.55 \\ 
 \setrow{\bfseries}
 &\normalfont{\textsc{Transformer}} (${\blacktriangleright}$) & 12.660\wideslash3.8920\wideslash0.8406\wideslash0.2577 & 0.7859 & 4.562 & 10.950 & 37.42 & 28.95 & 64.02 & 41.96 & 6.918\wideslash10.80\\ \cline{2-11}
 &HRAN & 12.190\wideslash3.8290\wideslash0.7752\wideslash0.2171 & 0.5074 & 2.883 & 7.104  & 36.53 & 28.08 & 63.58 & 40.22 & {\textbf{6.964}}\wideslash10.83\\
 \setrow{\bfseries} 
 &\normalfont{HRAN} (${\blacktriangleright}$) & 13.430\wideslash4.5030\wideslash1.0630\wideslash0.3513 & 0.7713 & 4.974 & 12.380 & 39.04 & 30.08 & 64.48 & 46.63 & {\normalfont{6.942}}\wideslash10.87 \\ 
 \midrule
 \multirow{8}{*}{(b)} 
 & \textsc{Seq2Seq} & 5.585\wideslash0.7887\wideslash0.1008\wideslash0.0296 & 1.1610 & 6.105 & 13.100 & 46.75 & 36.80 & \textbf{53.52} & 52.13 & 7.225\wideslash11.13 \\
 \setrow{\bfseries} 
 &\normalfont{\textsc{Seq2Seq}} (${\blacktriangleright}$) & 5.821\wideslash0.7910\wideslash0.1053\wideslash0.0377 & 1.3010 & 7.935 & 18.070 & 46.96 & 36.99 & \normalfont{53.41} & 53.40 & 7.464\wideslash11.66  \\ \cline{2-11}
 & HRED & {\textbf{5.899}}\wideslash0.7925\wideslash0.0786\wideslash0.0206 & 0.8334 & 5.147 & 14.160 & 48.12 & 36.50 & \textbf{54.20} & 49.99 & 7.107\wideslash10.90 \\
 \setrow{\bfseries} 
 & \normalfont{HRED} (${\blacktriangleright}$) 
 & {\normalfont{5.778}}\wideslash0.7968\wideslash0.0996\wideslash0.0387 & 1.2910 & 7.461 & 19.450 & 48.23 & 36.51 & {\normalfont{53.34}} & 50.31 & 7.436\wideslash11.10  \\ 
 \cline{2-11}
 & \textsc{Transformer} & 5.229\wideslash0.6443\wideslash0.0764\wideslash0.0240 & 1.1140 & 5.658 & 13.830 & 45.45 & 35.45 & {{53.04}} & 48.04 & 7.084\wideslash{{11.15}}  \\ 
 \setrow{\bfseries}
 & \normalfont{\textsc{Transformer}} (${\blacktriangleright}$) & 5.386\wideslash0.6460\wideslash0.0889\wideslash0.0274 & 1.3280 & 6.723 & 15.800 & 45.96 & 36.11 & 53.33 & 48.92 & 7.238\wideslash11.16  \\ \cline{2-11}
 & HRAN & 5.366\wideslash0.7229\wideslash{\textbf{0.0860}}\wideslash0.0182 & 1.0960 & 6.679 & 17.250 & 47.44 & 36.35 & \textbf{53.93} & 50.25 & 7.202\wideslash{\textbf{11.15}}  \\
 \setrow{\bfseries} 
 & \normalfont{HRAN} (${\blacktriangleright}$) & 5.541\wideslash0.7424\wideslash{\normalfont{0.0723}}\wideslash0.0194 & 1.6630 & 10.030 & 24.240 & 48.01 & 36.99 & \normalfont{53.46} & 51.81 & 7.394\wideslash{\normalfont{10.94}} \\
 \midrule
 \multirow{8}{*}{(c)} 
 & \textsc{Seq2Seq} & 5.666\wideslash1.0870\wideslash{\textbf{0.2471}}\wideslash0.0416 & 0.2880 & 2.110 & 5.566 & 54.22 & 46.11 & 63.96 & 56.82 & 6.685\wideslash10.54  \\
 \setrow{\bfseries} 
 & \normalfont{\textsc{Seq2Seq}} (${\blacktriangleright}$) & 5.696\wideslash1.1290\wideslash{\normalfont{0.2199}}\wideslash0.0476 & 0.4495 & 3.681 & 10.860 &  54.32 & 47.13 & 64.54 & 58.60 & 6.792\wideslash10.80\\ \cline{2-11}
 & HRED & 5.489\wideslash0.9953\wideslash0.2206\wideslash0.0711 & 0.3020 & 2.179 & 6.355 & \textbf{54.61} & \textbf{54.36} & 67.91 & 56.45 & 6.699\wideslash10.74 \\
 \setrow{\bfseries} 
 & \normalfont{HRED} (${\blacktriangleright}$) & 5.670\wideslash1.0930\wideslash0.2461\wideslash0.0828 & 0.4490 & 3.099 & 8.949 &  \normalfont{54.19} & 54.36 & 68.16 & 57.26 & 6.722\wideslash10.80 \\ \cline{2-11}
 & \textsc{Transformer} & 4.619\wideslash\textbf{0.8294}\wideslash0.1500\wideslash0.0307 & 0.3470 & 2.038 & 5.028 & 52.29 & 44.21 & 63.16 & 53.40 & 6.677\wideslash10.40 \\ 
 \setrow{\bfseries}
 & \normalfont{\textsc{Transformer}} (${\blacktriangleright}$) & 4.712\wideslash{\normalfont{0.8197}}\wideslash0.1744\wideslash0.0314 & 0.3897 & 2.437 & 6.188 &  52.34 & 45.12 & 63.52 & 54.11 & 6.722\wideslash10.50 \\ \cline{2-11}
 & HRAN & 5.090\wideslash0.8424\wideslash0.1665\wideslash0.0405 & 0.3205 & 2.604 & 8.188 & \textbf{54.74} & 54.52 & 68.16 & 56.58 & 6.556\wideslash10.53 \\
 \setrow{\bfseries} 
 & \normalfont{HRAN} (${\blacktriangleright}$) & 5.423\wideslash0.9192\wideslash0.1913\wideslash0.0529 & 0.5034 & 3.935 & 11.920 & \normalfont{54.40} & 54.54 & 68.30 & 57.48 & 6.699\wideslash10.89 \\
\bottomrule
\end{tabular}
}
\caption{
    Automatic evaluation results (\%) on the test set of three datasets: (a) PersonaChat, (b) Douban and (c) OpenSubtitles. 
    ``${\blacktriangleright}$'' denotes that the model is trained using our proposed framework.
    {The metrics Average, Extrema, Greedy and Coherence are abbreviated as Avg., Ext., Gre. and Coh., respectively.}
    The best results in each group are highlighted with \textbf{bold}.
}
\label{tbl:main_res}
\end{table*}

\section{Experiments}

\subsection{Experiment Settings}
\paragraph{Datasets}
We perform experiments on three conversation datasets: PersonaChat~\citep{DBLP:conf/acl/KielaWZDUS18}, Douban Corpus~\citep{DBLP:conf/acl/WuWXZL17} and OpenSubtitles~\citep{DBLP:conf/lrec/LisonT16}.
\textbf{PersonaChat}, an English-language dataset, contains multi-turn dialogues between pairs of speakers, collected via Amazon Mechanical Turk.
\textbf{Douban} consists of daily conversations from a popular social networking service---Douban group\footnote{\url{https://www.douban.com/group}} in China.
\textbf{OpenSubtitles} contains human-human conversations converted from movie transcripts in English.
Data statistics are listed in Table~\ref{tbl:data_statics}.

\paragraph{Experimental Models}
We apply the proposed group-wise contrastive learning framework to several state-of-the-art models, including
(\rom{1}) \textbf{\textsc{Seq2Seq}}: a LSTM-based sequence-to-sequence model with attention mechanisms~\citep{Bahdanau2014NeuralMT},
(\rom{2}) \textbf{HRED}: a hierarchical recurrent neural dialogue generation model~\citep{DBLP:conf/aaai/SerbanSBCP16},
(\rom{3}) \textbf{\textsc{Transformer}}: an encoder-decoder architecture relying solely on attention mechanisms~\citep{DBLP:conf/nips/VaswaniSPUJGKP17},
(\rom{4}) \textbf{HRAN}: a hierarchical recurrent attention network for multi-turn response generation ~\citep{DBLP:conf/aaai/XingWWHZ18}. 
Each model is trained using two protocols: 
the vanilla MLE training procedure and the proposed group-wise contrastive learning procedure, keeping other configurations the same.

\paragraph{Baselines}
We compare our group-wise contrastive learning framework against the following dialogue learning approaches:
(\rom{1}) \textbf{\textsc{Adversarial}}: an adversarial training approach for response generation~\citep{DBLP:conf/emnlp/LiMSJRJ17},
(\rom{2}) \textbf{MMI}: a training objective which maximums the mutual information between the dialogue context and its response~\citep{DBLP:conf/naacl/LiGBGD16,DBLP:conf/nips/ZhangGGGLBD18},
(\rom{3}) \textbf{\textsc{DeepRL}}: a reinforcement learning framework for neural response generation with heuristic reward functions to boost response qualities~\citep{DBLP:conf/emnlp/LiMRJGG16},
(\rom{4}) \textbf{CVAE}: a conditional variational auto-encoder learning framework to maximize the data likelihood, augmented with the KL-annealing technique~\citep{bowman-etal-2016-generating} and a BOW loss~\citep{DBLP:conf/acl/ZhaoZE17},
and (\rom{5}) \textbf{\textsc{DialogWAE}}: a conditional Wasserstein auto-encoder framework, modeling the distribution of dialogues by training a GAN within the latent variable space~\citep{DBLP:conf/iclr/GuCHK19}.

\paragraph{Automatic Evaluation Metrics}
We adopt several standard metrics widely used in existing works to measure the performance of dialogue generation models,
including BLEU~\citep{DBLP:conf/acl/PapineniRWZ02}, embedding-based metrics {(Average, Extrema, Greedy and Coherence)}\footnote{\url{https://chateval.org/}}~\citep{DBLP:conf/emnlp/LiuLSNCP16,DBLP:conf/emnlp/XuDKR18,N19-4011}, entropy-based metrics\footnote{We compute the entropy value for the empirical distribution of n-grams based on the maximum likelihood estimation on the training data.} {(Ent-\{1,2\})}~\citep{DBLP:journals/corr/SerbanSLCPCB16} and distinct metrics (Dist-\{1,2,3\})~\citep{DBLP:conf/naacl/LiGBGD16}.

\subsection{Implementation and Reproducibility}

We implement our model in ParlAI~\citep{miller-etal-2017-parlai} and train them on Nvidia P40 GPUs.
All the models use pretrained word embeddings produced by fastText~\citep{bojanowski2016enriching}, and the dimensionality of word vectors is 300.
For experimental models, 2-layer LSTM-based encoder and decoder with hidden size 256 are used in \textsc{Seq2Seq}.
We use the base Transformer configuration described in ~\citet{DBLP:conf/nips/VaswaniSPUJGKP17}, i.e., 6 blocks with 8 attention heads and 512 hidden units.
2-layer GRU-based RNNs are employed to build the word-level encoder, utterance-level encoder and response decoder for both the HRED and HRAN.
The GRU hidden size is set to 256.
For both models using different training procedures, we pretrain them by MLE and the result checkpoint is adopted as the reference model used in our framework.
We employ BM25~\citep{DBLP:journals/ftir/RobertsonZ09} to construct the index used during the contrastive dual sampling procedure.
The group size $\bm{k}$ is set to 3.
Regarding comparison models, we adopt the default configurations used in the original papers.
We optimize models by Adam~\citep{DBLP:journals/corr/KingmaB14} with an initial learning rate
of 0.001 and the batch size of 128.
All systems are trained until the validation loss fails to decrease for 5 checkpoints.
We compute the loss on the validation set at every 0.5 epochs and save the parameters for the top model.
Evaluation scores on the test set from the saved model are finally reported.

\subsection{Evaluation Results}
\paragraph{Performance on Experimental Models}
We instantiate the proposed framework on several state-of-the-art models for dialogue generation. 
Table~\ref{tbl:main_res} reports the automatic evaluation results of our learning framework and the conventional MLE training procedure.
By training dialogue models using the proposed learning framework, we witness solid performance boosts on three conversation datasets in terms of almost all the evaluation metrics, compared to the vanilla training.
Such improvements are also consistent across various experimental architectures, affirming the general applicability and superiority of the proposed framework.

\begin{table*}[!ht]
\centering
\resizebox{1.0\textwidth}{!}{
\centering
\begin{tabular}{>{\rowmac}l>{\rowmac}r>{\rowmac}r>{\rowmac}r>{\rowmac}r>{\rowmac}r>{\rowmac}r>{\rowmac}r>{\rowmac}r>{\rowmac}r<{\clearrow}}
\toprule
  Learning Approaches & \centercell{BLEU-1\wideslash2\wideslash3\wideslash4} & \centercell{Dist-1} & \centercell{Dist-2} & \centercell{Dist-3} & \centercell{Avg.} & \centercell{Ext.} & \centercell{Gre.} & \centercell{Coh.} & \centercell{Ent-1\wideslash2} \\
 \midrule
 \textsc{Adversarial} & 12.190\wideslash4.0060\wideslash0.8950\wideslash0.2644 & 0.6269 & 3.357 & 7.374 & 35.93 & 29.00 & 63.65 & 42.38 & 6.980\wideslash10.88 \\
 MMI & {\textbf{14.030}}\wideslash4.6460\wideslash1.3340\wideslash0.5022 & 0.4734 & 2.443 & 5.515 & 39.34 & 30.92 & {\textbf{64.84}} & 45.16 & 6.874\wideslash10.65 \\
 \textsc{DeepRL} & 12.660\wideslash4.0150\wideslash1.0140\wideslash0.3314 & 0.6838 & 3.838 & 8.581 & 37.23 & 29.68 & 64.30 & 44.13 & 6.885\wideslash10.85 \\
 CVAE & 11.570\wideslash2.8100\wideslash0.6357\wideslash0.1714 & 0.2876 & 2.326 & 7.506 & 39.29 & 30.61 & 63.67 & 41.76 & 6.869\wideslash10.82 \\ 
 \textsc{DialogWAE} & 11.430\wideslash2.9260\wideslash0.5676\wideslash0.1436 & \textbf{0.9936} & 5.080 & 9.928 & 38.68 &	28.70 & 63.39 & 41.06 & 7.009\wideslash{\textbf{11.09}} \\
 \hline
 \setrow{\bfseries} 
 Ours & {\normalfont{13.660}}\wideslash4.9160\wideslash1.5970\wideslash0.6122 & \normalfont{0.8492} & 5.093 & 12.000 & 39.76 & 31.74 & {\normalfont{64.76}} & 49.39 & 7.016\wideslash{\normalfont{10.90}} \\
\bottomrule
\end{tabular}
}
\caption{
    Performance (\%) of our approach instantiated on naive \textsc{Seq2Seq} and baseline approaches on PersonaChat.
}
\label{tbl:compare_baselines}
\end{table*}
\begin{table}[!t]
    \centering
    \resizebox{0.90\columnwidth}{!}{
    \centering
    \begin{tabular}{l>{\rowmac}c>{\rowmac}>{\rowmac}c>{\rowmac}c>{\rowmac}c<{\clearrow}}
        \toprule
         \textbf{Opponent}   & \textbf{Win} & \textbf{Loss} & \textbf{Tie} & \textbf{Kappa}  \\ \midrule
         Ours \textit{vs.} \textsc{Vanilla} MLE    & 53\% & 10\% & 37\% & 0.5750  \\ 
         Ours \textit{vs.} \textsc{Adversarial}    & 47\% & 15\% & 38\% & 0.5495  \\
         Ours \textit{vs.} MMI                     & 43\% & 12\% & 45\% & 0.5863  \\
         Ours \textit{vs.} \textsc{DeepRL}         & 40\% & 22\% & 38\% & 0.6036  \\  
         Ours \textit{vs.} CVAE                    & 40\% & 15\% & 45\% & 0.5510  \\
         Ours \textit{vs.} \textsc{DialogWAE}      & 45\% & 18\% & 37\% & 0.4216  \\
        \bottomrule 
    \end{tabular}
    }
    \caption{The results of human evaluation on the test set of PersonaChat.}
    \label{tbl:human_eval}
\end{table}
\begin{table*}[!t]
\centering
\resizebox{1.0\textwidth}{!}{
\centering
\begin{tabular}{>{\rowmac}l>{\rowmac}r>{\rowmac}r>{\rowmac}r>{\rowmac}r>{\rowmac}r>{\rowmac}r>{\rowmac}r>{\rowmac}r>{\rowmac}r<{\clearrow}}
\toprule
  Framework variants & \centercell{BLEU-1\wideslash2\wideslash3\wideslash4} & \centercell{Dist-1} & \centercell{Dist-2} & \centercell{Dist-3} & \centercell{Avg.} & \centercell{Ext.} & \centercell{Gre.} & \centercell{Coh.} & \centercell{Ent-1\wideslash2} \\
 \midrule
 (a) \textit{w/o} group-wise sampling             & 12.870\wideslash4.102\wideslash0.9564\wideslash0.2308 & 0.3965 & 2.070 & 4.633  & 36.52 & 29.09 & 64.21 & 42.40 & 6.836\wideslash10.62 \\ \hline
 (b) \textit{w/o} group-wise positive sampling    & 13.120\wideslash4.800\wideslash1.4180\wideslash0.5967 & 0.4632 & 2.270 & 5.002 & 38.26 & 31.18 & 64.66 & 43.03 & 6.812\wideslash10.49 \\
 (c) \textit{w/o} group-wise negative sampling    & 13.210\wideslash4.698\wideslash1.3970\wideslash0.5587 & 0.7175 & 3.532 & 7.473 & 38.23 & 30.96 & 64.62 & 46.27 & 6.911\wideslash10.68 \\ \hline
 (d) \textit{w/o} response-side sampling    & 13.340\wideslash4.730\wideslash1.4820\wideslash0.5779 & 0.8487 & 4.964 & 11.340 & 39.31 & 31.51 & 64.66 & 48.35 & 6.938\wideslash10.75\\
 (e) \textit{w/o} context-side sampling   & 13.170\wideslash4.539\wideslash1.4160\wideslash0.5308 & 0.8455 & 4.892 & 11.210 & 39.57 & \textbf{31.81} & 64.56 & 47.19 & 6.904\wideslash10.66 \\ \hline
 (f) \textit{w/o} impact of matching scores  & 13.560\wideslash4.359\wideslash1.1140\wideslash0.3823 & 0.6086 & 3.809 & 9.037 & 38.78 & 30.35 & 64.44 & 46.88 & 6.952\wideslash{\textbf{10.90}} \\
 \hline
 \setrow{\bfseries} 
 Full version & {{13.660}}\wideslash4.916\wideslash1.5970\wideslash0.6122 & 0.8492 & 5.093 & 12.000 & 39.76 & \normalfont{31.74} & {{64.76}} & 49.39 & 7.016\wideslash{{10.90}} \\
\bottomrule
\end{tabular}
}
\caption{
    Ablation test (\%) using \textsc{Seq2Seq} with different framework variants on PersonaChat.
}
\label{tbl:ablation}
\end{table*}

\paragraph{Comparison with Baseline Approaches}
We compare our proposed framework with existing learning approaches designed for dialogue generation task.
Table~\ref{tbl:compare_baselines} summarizes the evaluation results.
We observe that our learning framework outperforms previous approaches regarding the majority of evaluation metrics.
It is worth noting that the proposed framework brings a relatively large improvement regarding both the response diversity and conversation coherence, indicating that our approach helps the dialogue model to generate not only informative but also context-relevant responses, which confirms our hypothesis that the group-wise contrastive learning encourages distinctiveness.

\paragraph{Human Evaluation}
We further conduct human evaluations to assess the proposed learning framework.
We choose the PersonaChat as our evaluation corpus since its expressions are more close to the style of daily chat and are easier for the annotators to make judgments.
Three crowd-sourced graduate students are employed to evaluate the quality of generated responses for 100 randomly sampled input contexts.
During the evaluation, the annotators are requested to select a preferred response, or vote a tie, considering the following aspects of response quality: fluency, informativeness, coherence and engagingness.
Table~\ref{tbl:human_eval} summarizes the evaluation results and the Cohen's kappa scores~\citep{cohen1960coefficient} to measure the intra-rater reliability.
We observe that our learning framework brings more preferable replies compared with the competitors.
This indicates that training the dialogue model with the proposed group-wise contrastive learning framework does improve the response quality.

\subsection{Model Analysis}

\paragraph{Effect of the Group-wise Learning Strategy}

To manage the multi-mapping relations in human-human conversation and stabilize the model training with noisy data, the dialogue model is induced to contrast a group of positive samples from the negative ones, pulling the matched sample pairs together and pushing the mismatched pairs apart.
We ablate the group-wise learning from the framework by using only one pair of positive and negative samples to verify its effectiveness.
As presented in Table~\ref{tbl:ablation} (a), we can see that disabling group-wise learning hurts performance significantly on all evaluation metrics.
Note that ablating either the group-wise positive sampling (Table~\ref{tbl:ablation} (b)) or group-wise negative sampling (Table~\ref{tbl:ablation} (c)) also leads to a performance drop with respect to the evaluation metrics.
It demonstrates that the group-wise learning strategy plays a key role in achieving strong performance.

\paragraph{Effect of the Dual Sampling}
In our framework, the contrastive samples can be organized regarding either the dialogue context or response, allowing the dialogue model to explore both the many-to-one and one-to-many relations in human conversation. 
We investigate different sampling strategies in Table~\ref{tbl:ablation} (d) and (e).
We notice that when both the response-side and context-side samplings together incorporated into the learning framework, the model achieves its best performance, verifying the effectiveness of the contrastive dual sampling.

\paragraph{Impact of Matching Scores}
To discriminatively exploit the sampled context-response pairs with varied matching degrees, we utilize the matching score attached with each sample to adapt its instance effect on the model training.
We conduct the ablation test of this learning strategy by simply discarding the impact of matching scores as in Eq.(\ref{eq:loss'}).
As shown in Table~\ref{tbl:ablation} (f), training without considering matching degrees of samples leads to a consistent performance drop, which suggests that the system can benefit from perceiving fine-grained differences within the group during contrastive learning.

\begin{figure*}[!thp]
    \centering 
    \includegraphics[width=0.93\textwidth]{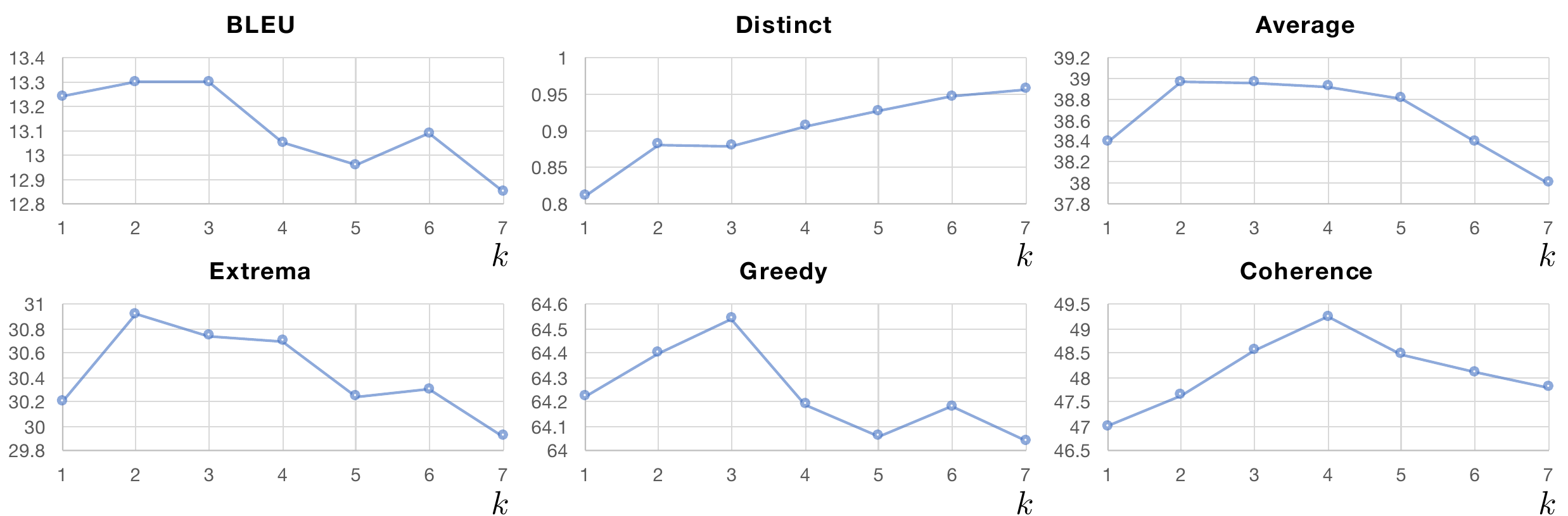}
    \caption{
        Evaluation results (\%) with different group size $\bm{k}$ on the validation set of PersonaChat using the proposed framework instantiated on \textsc{Seq2Seq}. BLEU-1 and Dist-1 are denoted as ``BLEU'' and ``Distinct'', respectively.
    }
    \label{fig:ablation_k}
\end{figure*}
\paragraph{Impact of Group Size}
We explore the impact of using different group size $\bm{k}$ in our group-wise contrastive learning framework in Figure~\ref{fig:ablation_k}.
We observe that increasing the group size $\bm{k}$ leads to continuous improvement on the Distinct metric while other reference-based metrics achieve the best results at a moderate group size.
We conjecture that a larger group size allows the dialogue model to learn from more diverse expressions, meanwhile it also risks introducing more utterances that are inconsistent with the references.

\section{Related Work}
\paragraph{Learning Methods for Dialogue Generation}
Typically, state-of-the-art neural dialogue generation models adopt \textit{Maximum Likelihood Estimation (MLE)} as their learning approach, maximizing log-likelihood of model parameters on the training data. 
Though effective, well-known issues, including the notorious general dull response problem, are reported by prior art~\citep{DBLP:conf/naacl/LiGBGD16,DBLP:conf/acl/ZhaoZE17} on dialogue models trained with MLE.

Alternative dialogue learning approaches are proposed to tackle the issues.
\citet{DBLP:conf/naacl/LiGBGD16,DBLP:conf/nips/ZhangGGGLBD18} introduce the Maximum Mutual Information as the objective function to promote response diversity.
Techniques of reinforcement learning (RL) and adversarial learning have been introduced into dialogue generation by ~\citet{DBLP:conf/emnlp/LiMRJGG16,DBLP:conf/emnlp/LiMSJRJ17} to better approximate the optimization goal of dialogue models.
Conditional variational framework~\citep{DBLP:conf/acl/ZhaoZE17,DBLP:conf/acl/ShenSLLNZAL17,park-etal-2018-hierarchical} has also shown a promise in dialogue generation.
\citet{DBLP:conf/iclr/GuCHK19} further introduce a conditional Wasserstein autoencoder that employs GAN~\citep{NIPS2014_5423} to model the multimodal latent structures.
\citet{cai-etal-2020-curricula} design a multi-curriculum learning framework to facilitate the dialogue model training.
\checkhere{Contrasted with existing learning methods for dialogue generation, the proposed framework in this work encourages the model to learn from the difference between well-chosen contrastive pairs, which explicitly models the multi-mapping relations in conversation and promotes the distinctiveness of the generated responses in the meantime.}

\paragraph{Contrastive Learning}
The concept of learning by contrasting positive pairs against negative pairs~\citep{DBLP:conf/cvpr/HadsellCL06,DBLP:journals/jmlr/GutmannH12} has been successfully adopted in many tasks.
For example, contrastive learning in language modeling task~\citep{10.5555/3042573.3042630,DBLP:conf/emnlp/VaswaniZFC13,DBLP:conf/naacl/BaltescuB15} aims to approximate the negative log-likelihood by training the model to correctly classify between generated noise samples and words observed in the training data.
Contrastive visual representation learning~\citep{DBLP:journals/corr/abs-1807-03748,DBLP:journals/corr/abs-2002-05709} trains a generative model to score real data points higher than negative samples.
\citet{CL4ImageCaption} propose to use contrastive learning for image caption.
\citet{DBLP:conf/iclr/ClarkLLM20} use contrastive learning to train a discriminative model for language representation learning.
\checkhere{Compared with existing work, samples used in this paper, instead of being sampled randomly, are carefully chosen to exhibit particular properties of human dialogues.
Another difference is that, we manage multi-mapping relations prevalent in human conversation using many positives and many negatives, which captures both the intra-group and inter-group variability.}

\section{Conclusion}

In this work, we propose a group-wise contrastive dialogue learning approach, that explicitly perceives the difference between the well-chosen positive and negative utterances, and manages the multi-mapping relations in human conversations simultaneously. 
Given a training instance, the proposed learning framework first organizes a group of positive samples and negative samples regarding context-response matching degrees, and then trains the target dialogue model to give higher conditional probabilities for positive pairs and lower probabilities for the negatives.
Extensive experimental results show that the proposed learning framework brings a solid favorable performance boost amongst various strong baseline approaches.

\section*{Acknowledgments}
We would like to thank all the reviewers for their insightful and valuable comments and suggestions.
Hongshen Chen and Yonghao Song are the corresponding authors.

\bibliography{Main}
\bibliographystyle{emnlp2020-templates/acl_natbib}

\end{document}